\pdfoutput=1

\documentclass[11pt]{article}

\usepackage{acl}

\usepackage{times}
\usepackage{latexsym}

\usepackage[T1]{fontenc}

\usepackage[utf8]{inputenc}

\usepackage{microtype}

\usepackage{inconsolata}

\usepackage{graphicx}

\usepackage{graphicx}
\usepackage{enumerate}
\usepackage{amsmath}
\usepackage{multirow}

\usepackage{tcolorbox}

\usepackage{xcolor}
\usepackage{float}

\title{PDC \& DM-SFT: A Road for LLM SQL Bug-Fix Enhancing}

\author{Yiwen Duan, Yonghong Yu, Xiaoming Zhao, Yichang Wu, Wenbo Liu \\
  Bytedance Inc.  \\
  \texttt{[duanyiwen.86, yuyonghong, zhaoxiaoming.chuck, wuyichang, liuwenbo.arthur]@bytedance.com} 
  }

\begin{document}
\maketitle
\begin{abstract}
Code Large Language Models (Code LLMs), such as Code llama and DeepSeek-Coder, have demonstrated exceptional performance in the code generation tasks. However, most existing models focus on the abilities of generating correct code, but often struggle with bug repair.
We introduce a suit of methods to enhance LLM's SQL bug-fixing abilities. The methods are mainly consisted of two parts: A Progressive Dataset Construction (PDC) from scratch and Dynamic Mask Supervised Fine-tuning (DM-SFT). 
PDC proposes two data expansion methods from the perspectives of breadth first and depth first respectively. DM-SFT introduces an efficient bug-fixing supervised learning approach, which effectively reduce the total training steps and mitigate the "disorientation" in SQL code bug-fixing training. 
In our evaluation, the code LLM models trained with two methods have exceeds all current best performing model which size is much larger.
\end{abstract}

\section{Introduction}

Recently, as large language models (LLMs) achieve remarkable success, code LLMs emerge as useful assistants when editing code. However, when we shift focus to fixing code errors, we find that the performance of open source pre-trained code LLMs like DeepSeek-Coder \citep{guo2024deepseek}, WizardCoder \citep{luo2023wizardcoder} and Code Llama \citep{roziere2023code} is quite limited (as shown in Table~\ref{tab:masksftexp}).

In this paper, we especially focus on the code repair task of SQL. Due to the complex nested query structure, SQL code bugs are more difficult to solve compared with other code languages. We formulate the SQL code bug-fixing task as Equation~\ref{eq:sqlbugfix}.

\begin{equation}
  \label{eq:sqlbugfix}
  SQL_{correct} = f(\textcolor{blue}{Schema}, \textcolor{red}{SQL_{bug}}, \textcolor{cyan}{R})
\end{equation}

Where the $f$ represents the bug-fixing model. \textcolor{blue}{Schema} means the related tables schemas of bug SQL code. \textcolor{red}{SQL\textsubscript{bug}} denote the SQL code which contains some bugs need to be fixed. \textcolor{cyan}{R} is the return message by the SQL execution system when you run the bug SQL code. $SQL_{correct}$ is the bug-fixing model’s output, which is expected the right SQL code.

We propose a set of methods to enhance the bug-fixing capabilities of Large Language Models (LLMs). This includes a method for mining and collecting supervised data, termed Progressive Dataset Construction (PDC), and an efficient training method based on dynamic masking, known as Dynamic Mask-SFT (DM-SFT). Experiments show that training with data collected via PDC method generally improved the SQL bug-fixing capabilities of open-source code LLMs by nearly $+50\%$. The Dynamic Mask-SFT training method further enhanced model performance by approximately $+10\%$ relative to the default generative SFT.

\section{Related Work}

Deep learning-based code bug repair has attracted attention with the advancement of pre-trained LLMs. Most methods follow a zero/few-shot learning paradigm, directly using LLMs to generate repaired code from context. \citet{huang2023empirical} explored fine-tuning LLMs for bug fixing, showing significant improvements over previous tools.

Other approaches generate supervised data by transforming correct code into buggy code. BUGLAB \citep{allamanis2021self} uses self-supervision to train bug detectors, while Break-It-Fix-It \citep{yasunaga2021break} collaboratively trains bug fixers and generators. However, generating realistic SQL bugs remains challenging due to its differences from object-oriented languages.

Agent-based approaches like RepairAgent \citep{bouzenia2024repairagent} and SELF-DEBUGGING \citep{chen2023teaching} enable LLMs to autonomously fix bugs. But debugging SQL code at the task level is time-consuming, making repeated execution impractical.

\section{Progressive Dataset Construction}

\begin{figure}[t]
  \includegraphics[width=\columnwidth]{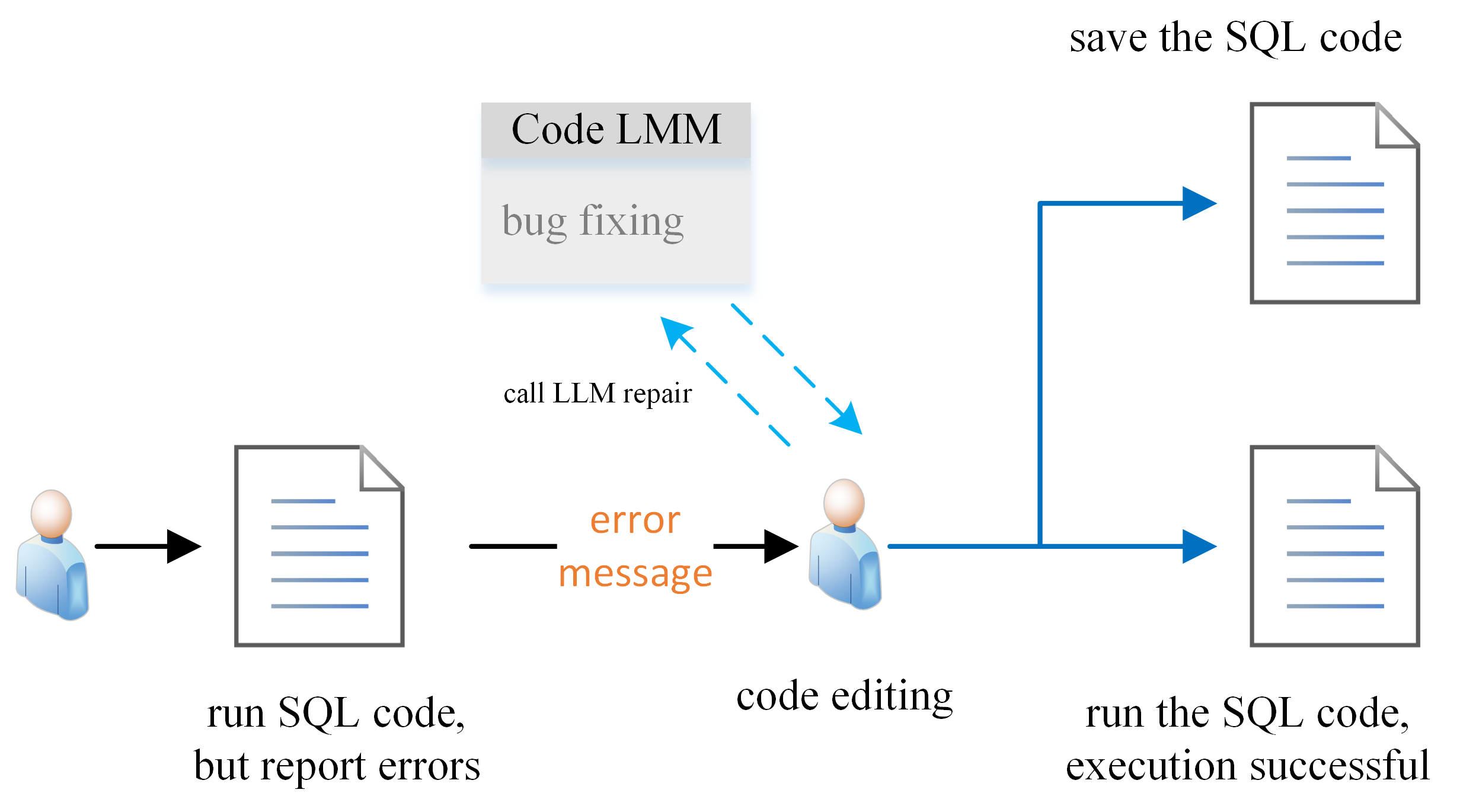}
  \caption{The initial training data collection via user behavior logs mining.}
  \label{fig:datacollect}
\end{figure}

In this section, we introduce a set of data collection methods called Progressive Dataset Construction (PDC). 
The methods include two parts: diverse collecting from online system (breadth first) and oriented generation of offline mining (depth first). 
The diverse collecting through automated methods ensures the diversity coverage and sustainable scalability of the training datasets, thereby maintaining a consistent alignment between the distribution of training data and the behaviors of online users. 
The oriented generation method is used for data augmentation in cases where the model performs poorly in evaluation and online serving. This approach requires assistance of code LLM and some SQL corpora recall methods.

\subsection{Diverse Collecting}

\textbf{Data Collecting}. To collect initial training data, we designed rules to mine online user behavior logs. As shown in Figure~\ref{fig:datacollect}, when users encounter SQL execution errors, the system logs the erroneous code and error message. Users then typically edit and correct the code until it runs successfully, allowing us to extract many $(bug SQL, correct SQL)$ pairs from their behavior.

Moreover, since the SQL environment includes syntax checking, some users modify their code based on syntax prompts and save it without re-executing when the highlighted syntax error prompts disappear. Thus, we also consider the last 'save code' operation after an execution error as a signal for identifying correct SQL, as depicted in Figure~\ref{fig:datacollect}.

\begin{figure}[t]
  \includegraphics[width=\columnwidth]{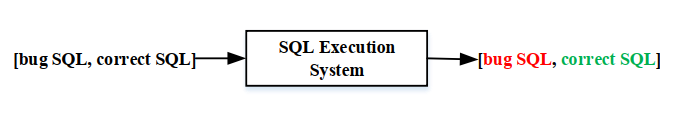}
  \caption{Execution filter for data quality.}
  \label{fig:datafilter}
\end{figure}

\textbf{Automated filtering}. After collecting data from online user logs, we apply an execution filter as shown in Figure~\ref{fig:datafilter}. This retains $(bug SQL, correct SQL)$ pairs where the bug SQL causes an error (red) and the correct SQL runs successfully (green). We also remove samples where the difference between the bug SQL and correct SQL is too large to ensure data quality.

\textbf{Spot check}. Lastly, we conduct a manual sampling inspection of the filtered data to ensure that correct SQL is just transformed from bug SQL by bug fix, without any SQL semantics change. If the [bug SQL, correct SQL] pairs achieve inspection pass rate over $85\%$, they meet our quality standards and are deemed suitable as training data. 

Diverse collecting samples for bug SQL repair directly from online user behavior ensures an excellent coverage of diversity. It aligns with the natural data distribution in real service scenarios, which is crucial for model training.
Even after model's serving online, diverse collection remains essential to identify cases where users reject model's fixes and make manual edits, indicating a mismatch with their expectations. 

\subsection{Oriented Generation}

Oriented generation is a data augmentation method targeting difficult cases, such as unique syntax features and rare long-tail error types. We used regex-based templates to classify bugs from error messages and codes, organizing them into 81 categories across three levels. As shown in the Appendix~\ref{sec:appfigures} Figure~\ref{appendix:bug_types}.

\begin{figure}[t]
  \includegraphics[width=\columnwidth]{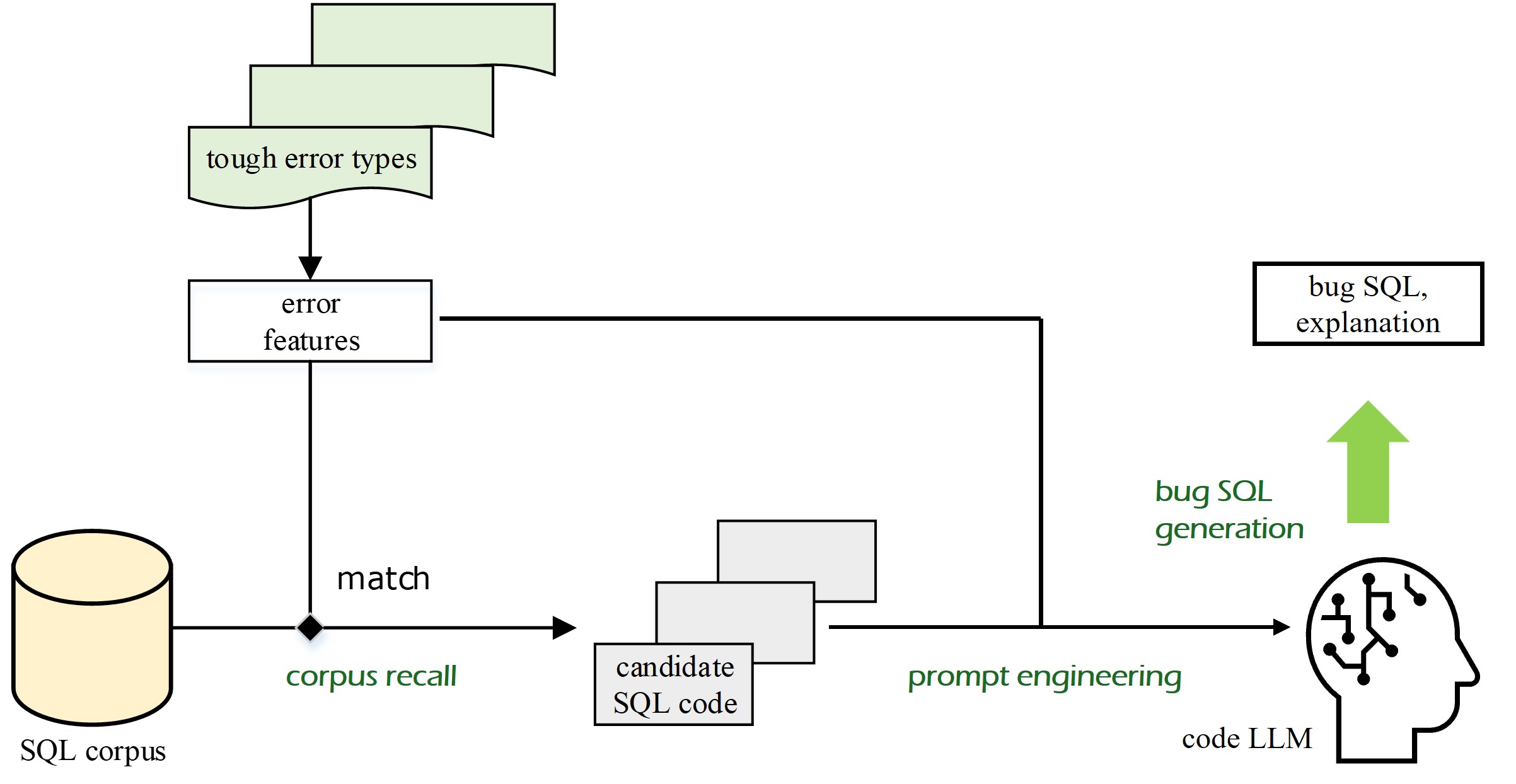}
  \caption{Overview of oriented generation method.}
  \label{fig:orientedgen}
\end{figure}

The original SQL corpus consists of executable SQL code from historical platform users. As illustrated in Figure~\ref{fig:orientedgen}, we apply this method to augment data for bug types that are challenging for the model, following the steps outlined below:
\begin{enumerate}[(1)]
        \item {\bf Identify target types.}
Initially, we target rare long-tail bugs. After deployment, we focus on types where model correction accuracy is low. 
        \item {\bf Define an “error feature” for each type.}
Error features depend on the recall algorithm used. For example, you can use syntax keywords for recall, such as using the keyword “group by” to match SQL code suitable for generating “group by” errors. 
        \item {\bf Recall candidate SQL code.}
We employ appropriate rule based matching algorithm to pair a rich corpus of SQL code with each bug type via "error feature". As accuracy of matching varies across different bug types, different matching algorithm for different bug type is needed sometimes. 
        \item {\bf Generate bug SQL samples for each bug type.}
This step requires assistance of a robust code LLM for the generation of bug SQL code. In our practice, the quality of bug SQL generated is closely tied to the prompt. We provide a reference prompt in Appendix~\ref{sec:appprompt} used in our internal code fundamental LLM for bug SQL generation.
\end{enumerate}

The diverse collecting and oriented generation methods respectively accomplish the supervised dataset construction for the SQL bug fixing task from the perspectives of breadth-first and depth-first approaches. Both methods remain effective post-deployment. The diverse collecting method, driven by user behavior, gathers unsatisfactory samples for improvement. Meanwhile, oriented generation can specifically enhance the types of bugs where the model's performance is subpar. The collected data can be utilized to improve the model's performance.  The enhancement of model performance, in turn, affects the distribution of the data collecting. Therefore, this is a progressive dataset construction method.

\section{Dynamic Mask Supervised Fine-tuning}
\begin{figure}[t]
  \includegraphics[width=\columnwidth]{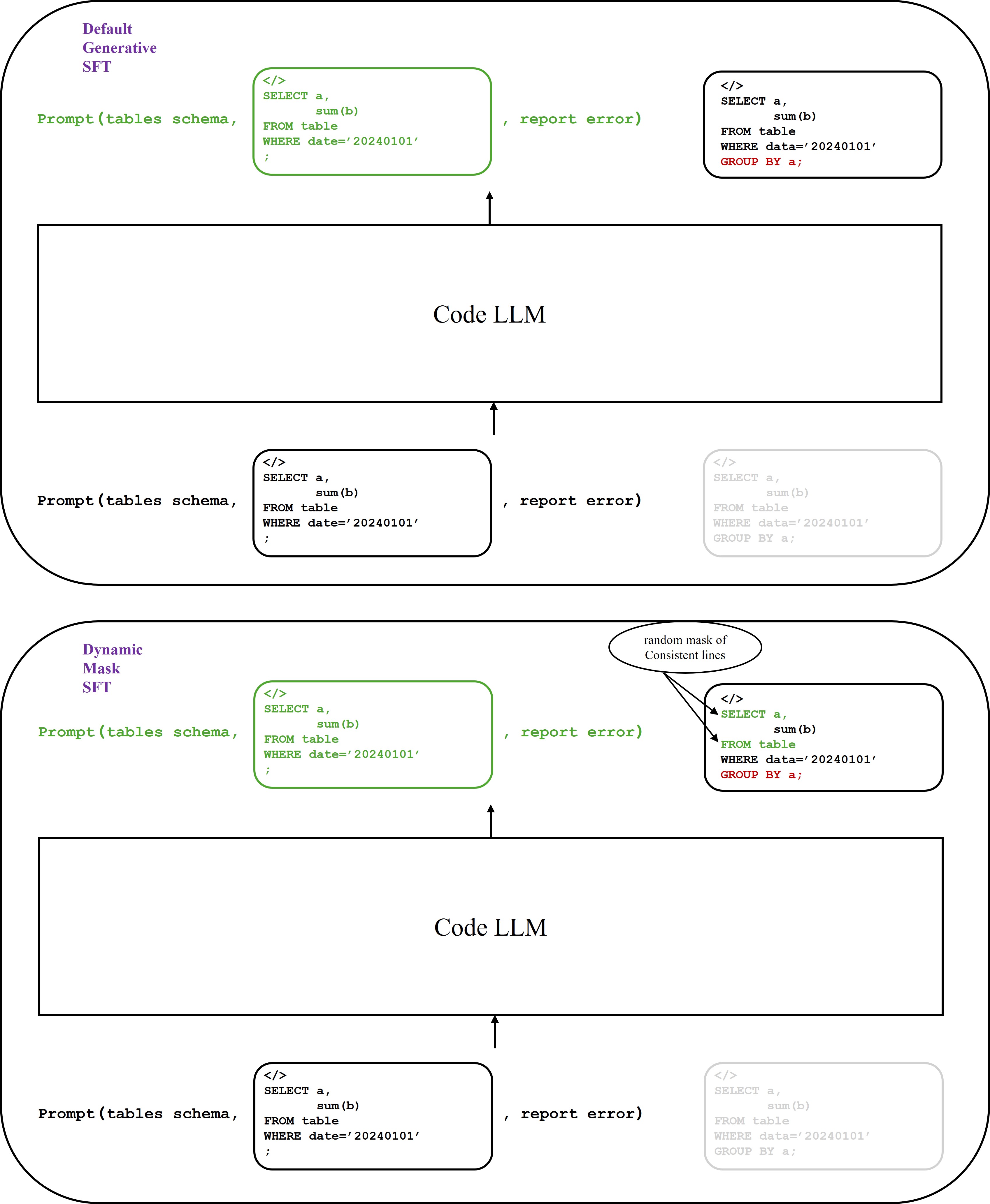}
  \caption{A comparison of the default generative SFT (top) and dynamic mask SFT (bottom) for the code bug-fixing task.}
  \label{fig:masksft}
\end{figure}

In this section, we present a detailed introduction to an efficient training method for LLM SQL code bug fixing, which refer to as dynamic mask supervised fine-tuning (DM-SFT). The Figure~\ref{fig:masksft} compares DM-SFT with default generative SFT in terms of training and loss calculation. As described in the Introduction, the input prompt (Appendix~\ref{sec:online_prompt}) composed of three pieces information: [tables DDL, bug SQL, report error]. The model's output is a complete, corrected SQL code. Notably, most lines between the bug SQL and correct SQL are identical, with only a few requiring changes.

In our collected training data, the count distribution of code lines that need to be modified when editing from bug SQL to correct SQL (called as diff lines) is shown in Appendix~\ref{sec:appfigures} Figure~\ref{appendix:difflines}. Over 92\% of cases have fewer than 5 diff lines, meaning most correct code is already present in the input (bug SQL). In default generative fine-tuning, all output tokens contribute equally to the calculation of final loss, leading to issues like slow convergence and unstable training, which we will detail in the experimental section.

To address these issues, we propose a code bug repair training method called dynamic mask SFT. During the model training process, we divide the correct SQL code that the model is expected to predict post bug-fixing into two categories in line-by-line basis:

\begin{enumerate}[(i)]
        \item {\bf Consistent lines}:
Code lines unchanged from the original bug SQL.  
        \item {\bf Diff lines}: 
Code lines that require modification. 
\end{enumerate}

Given a bug SQL code, related tables schema, report error and corresponding correct SQL code, we use $(l_0,l_1,l_2,\cdots,d_0,\cdots,d_m,\cdots l_n), m \leq n$ denoting the correct code lines. The $l_i,i \in
[0,n]$ represents the consistent lines and $d_j,j \in [0,m]$ represents the diff lines. We use $u$ to denote tokens of consistent lines, and $v$ to denote tokens of diff lines. Equation~\ref{eq:masksftloss} shows the loss function of dynamic mask SFT.

\begin{equation}
\label{eq:masksftloss}
  \begin{split}
  L_1 = & -\sum \log P(u_{k+1} \mid u_k,u_{k-1},\ldots,u_0) \\
  & \ast a(l(u_{k+1}))
  \end{split}
\end{equation}

\begin{equation}
\label{eq:maskweight}
   a(l_i) = 
  \begin{cases}
  0 & \text{p} \\
  1 & \text{(1-p)}
  \end{cases} 
\end{equation}

Where $a(l_i)$ is the mask weight of line $l_i$ as Equation~\ref{eq:maskweight}, and mask weight of all tokens in line $l_i$ are the same. The p is random mask ratio factor, used to control the proportion of masked code lines. $l(u_{k+1} )$ represents the line number of code where token $u_{k+1}$ is located. In Equation~\ref{eq:masksftloss}, $L_1$ represents the language model loss of the consistent lines (after dynamic masking). In Equation~\ref{eq:normaloss}, $L_2$ represents the language model loss of the diff lines.

\begin{equation}
\label{eq:normaloss}
  \begin{split}
  L_2 = & -\sum {log P(v_{k+1} \mid v_k,v_{k-1},\ldots,v_0) }
  \end{split}
\end{equation}

\begin{equation}
\label{eq:totalloss}
  L = L_1 + L_2
\end{equation}

The final total loss $L$, as shown in Equation~\ref{eq:totalloss}, is composed of $L_1$ and $L_2$.

Figure~\ref{fig:masksft} highlights the similarities and differences between dynamic mask SFT and default SFT on bug-fix training. The correct SQL code need to be predicted is shown in grey. In the output label, the parts that don't need to be calculated in loss are highlighted in green (input prompt and masked code lines randomly selected with probability \(p\)).

\section{Experiments and Results}

In this section, we present a detailed overview of our experimental setup and results, divided into two main parts. First, we demonstrate the effectiveness of the PDC and DM-SFT methods through a series of ablation experiments. Next, we analyze the impact of the random mask ratio \(p\) in DM-SFT on model performance and training efficiency. Finally, our analysis of hallucination issues found in model evaluation and experiments on reducing hallucinations through continue pre-train (CPT) ~\citep{ke2023continual} are detailed in Appendix~\ref{sec:cpt}.

\subsection{PDC and SFT Experiments}


We demonstrate the efficacy of the suit of methods (PDC \& DM-SFT) through a series of experiments. We collected 3k diverse samples through the diverse collecting method and 300+ oriented enhancement samples based on code LLM by the oriented generation method. Based on these 3.3k data\footnote{All SQL code was collected from the company's internal big data development platform and written by data engineers. Over $90\%$ of the SQL is task-level, with an average length more than 100 lines. These SQL scripts are highly diverse, covering various business scenarios such as e-commerce, short videos, search, and advertising.}, we conducted ablation experiments to verify DM-SFT's effectiveness.

\begin{table*}
  \centering
  \begin{tabular}{llcc}
    \hline
    \textbf{Method} & \textbf{Model} & \textbf{Size} & \textbf{Acc}\\
    \hline
    \multirow{9}{4em}{Pretrain} & gemma & 7B & $20.1\%$          \\
    & StarCoderBase                     & 7B & $20.8\%$          \\
    & StarCoder2                        & 7B & $22.5\%$          \\
    & CodeQwen1.5-Chat                  & 7B & $27.8\%$          \\
    & DeepSeek-Coder-instruct              & 6.7B & $28.5\%$          \\
    & DeepSeek-Coder-instruct              & 33B & $29.8\%$          \\
    & DeepSeek-Coder-V2-Lite-Instruct(MOE)      & 16B & $28.8\%$          \\
    & WizardCoder-V1.1                 & 33  B & $29.7\%$          \\
    & internal code LLM                    & * & $40.5\%$          \\
    \hline
    \multirow{6}{4em}{SFT} & gemma      & 7B & $29.0\%$          \\
    & StarCoderBase                     & 7B & $32.6\%$          \\
    & CodeQwen1.5-Chat                  & 7B & $42.6\%$          \\
    & DeepSeek-Coder-instruct              & 6.7B & $43.8\%$          \\
    & DeepSeek-Coder-V2-Lite-Instruct(MOE)      & 16B & $43.9\%$          \\
    & internal code LLM                    & * & $46.6\%$          \\
    \hline
    \multirow{3}{4em}{DM-SFT} & CodeQwen1.5-Chat & 7B & $49.3\%$          \\
    & DeepSeek-Coder-instruct             & 6.7B & $49.8\%$          \\
    & DeepSeek-Coder-V2-Lite-Instruct(MOE)     & 16B & $49.7\%$          \\
    \hline
  \end{tabular}
  \caption{ Accuracy of different models and training methods.}
  \label{tab:masksftexp}
\end{table*}

We use DeepSeek-Coder-instruct (6.7b)  as the fundamental model and carry out the training experiments on a cluster of 32 × NVIDIA A800 80GB GPUs using the DeepSpeed \citep{rajbhandari2020zero} framework stage 3. In terms of hyperparameters setting, we used batch size = 32, learning rate = 1.2e-5, and AdamW optimizer \citep{loshchilov2017decoupled} with $adam\_beta1=0.9$ and $adam\_beta2=0.95$ (more detailed experimental parameter configurations, please refer to code release information in the final part of this section).

We constructed a 1,072-entry evaluation dataset. 748 entries were randomly sampled from execution logs on in data platform, reflecting natural distribution of SQL error types in production. The remaining 324 entries were crafted to cover 81 error types(one type four examples). This ensures alignment with real-world scenarios and allows performance estimates on long-tail errors. The ground truth of the dataset is precisely annotated by staff SQL engineers. During the model development stage, we used machine automatic evaluation (a method based on AST semantic comparison) results to select the approximate best training steps and hyper-parameters. Besides, in some samples, there's more than one correct way to fix the bug. The final model's bug fixing accuracy was determined by human evaluation of staff SQL engineers.\footnote{Unlike typical data query platform SQL, the SQL code on a data development platform is often task-level, meaning a single execution can take several hours and incur high costs. In contrast, conducting reliable manual evaluations based on classified error type labels and ground truth is more practical and efficient.}

In the evaluation, we first assessed the bug-fixing capabilities of leading open code LLMs and our powerful internal code LLM (a closed-source code LLM, without any bug-fixing SFT enhancement) as a baseline to evaluate our PDC data collection methods. Furthermore, through ablation experiments, we compared the impact of dynamic mask SFT and default generative SFT on training. 

We conducted independent tests on various models, with outputs subjected to blind manual evaluation (evaluators were unaware of which model each answer came from, and each bug-fixing sample was cross reviewed by three individuals). The final fixing accuracy of each model on the 1072-sample evaluation dataset are shown in Table~\ref{tab:masksftexp}. 

It is evident that among the models with around the 7B parameters, DeepSeek-Coder-6.7B-instruct achieves the highest fixing accuracy. Additionally, we observe that the larger 33B model does not exhibit significant improvement compared to the 7B model. Using DeepSeek-Coder-6.7B-instruct as the foundational model, we conducted both default generative dynamic mask SFT training on the 3.3k training dataset collected through the PDC method.


As observed in Table~\ref{tab:masksftexp}, the 3.3k samples from the PDC method (Diverse collecting \& Oriented generation) significantly boosted the DeepSeek-Coder model's bug-fixing accuracy from $28.5\%$ to $43.8\%$, a relative increase of over $50\%$. We also conducted SFT experiments on other models with parameter sizes around 7B, DeepSeek-Coder-V2-Lite-Instruct \citep{zhu2024deepseek} and internal code LLM, the findings were consistent.

Furthermore, we employed dynamic mask SFT to train models on DeepSeek-Coder-6.7B-instruct, CodeQwen1.5-7B-Chat and DeepSeek-Coder-V2-Lite-Instruct, top-performing models in default SFT. Results from manual evaluations indicate that dynamic mask SFT can enhance the model's bug fixing capability by approximately $10\%$ compared to the default generative SFT training (DeepSeek-Coder-6.7B-instruct: $43.8\%$→$49.8\%$, CodeQwen1.5-7B-Chat: $42.6\%$→$49.3\%$), DeepSeek-Coder-V2-Lite-Instruct: $43.9\%$→$49.7\%$).

\subsection{Mask Ratio Experiments}

\begin{figure}[t]
  \includegraphics[width=\columnwidth]{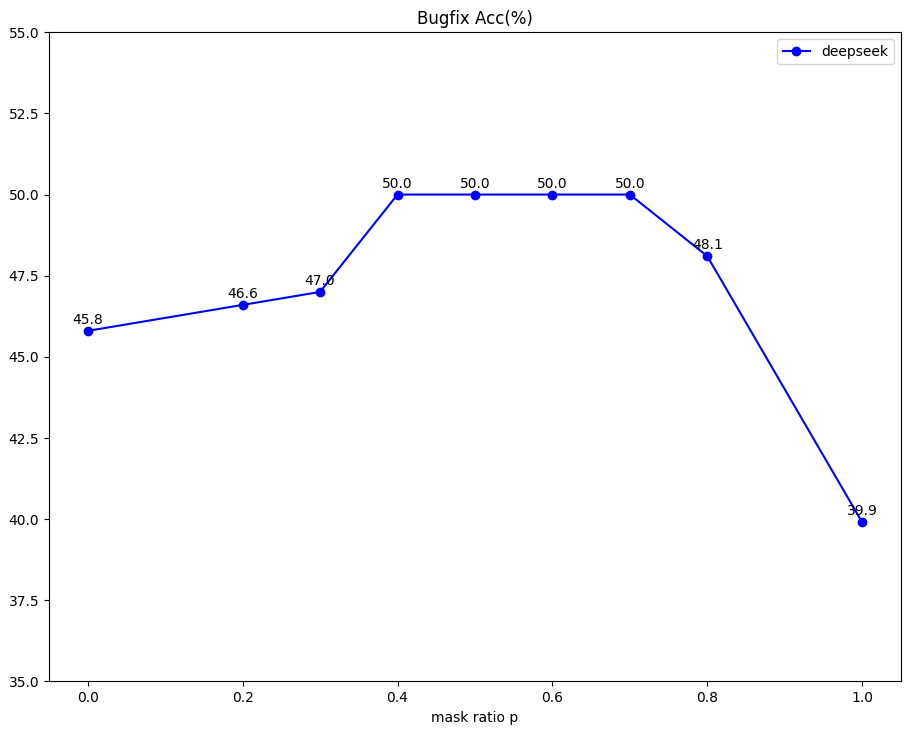}
  \caption{Bug fixing evaluation results with different value of random mask ratio factor $p$.}
  \label{fig:maskratio}
\end{figure}

\begin{figure}[t]
  \includegraphics[width=\columnwidth]{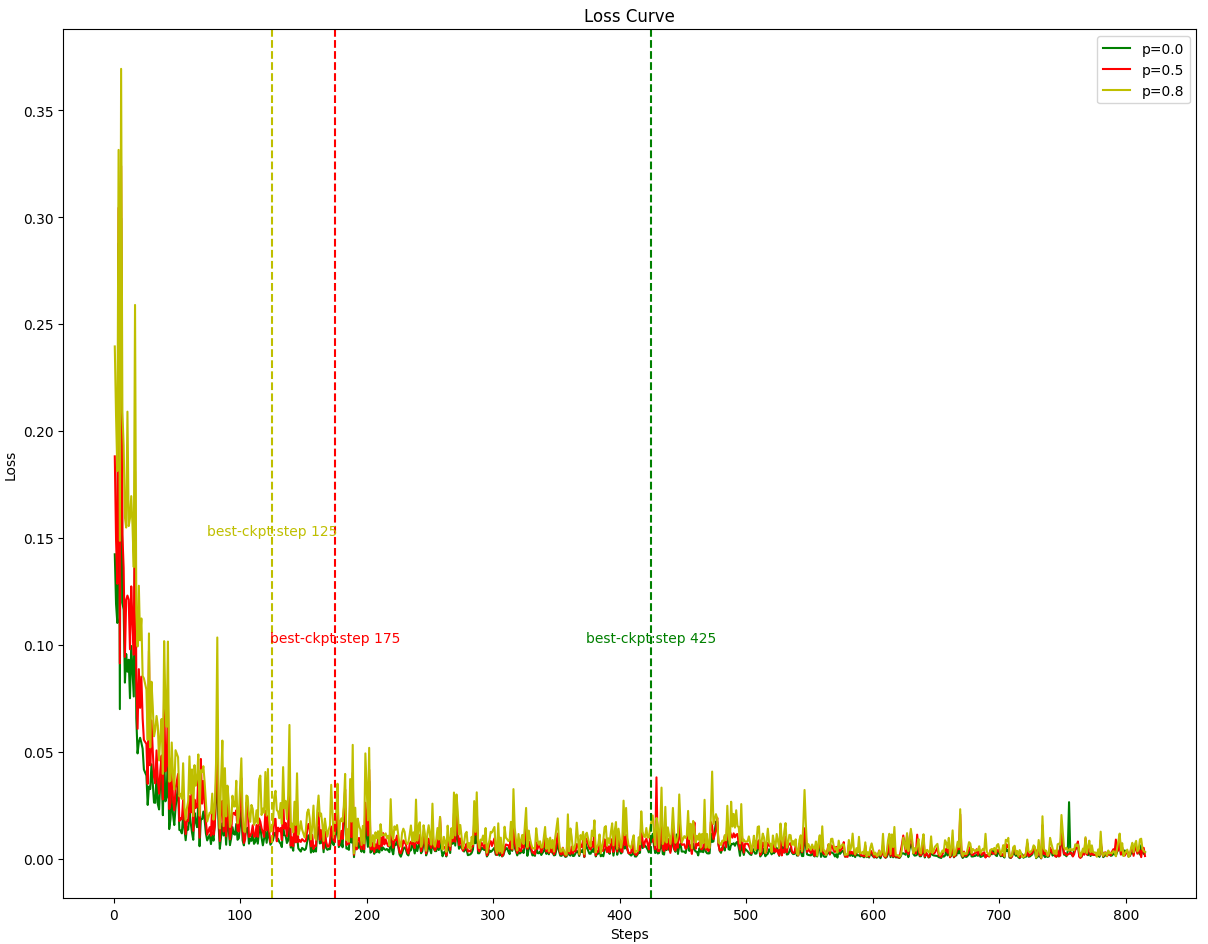}
  \caption{ Loss reduction curves and best bug fixing performance steps across typical random mask ratio factors \(p\) during model training.}
  \label{fig:loss}
\end{figure}

Taking the best-performing DeepSeek-Coder-6.7B-instruct as the foundation model, we trained with different \(p\) values to evaluate bug-fixing capability. The results presented in Figure~\ref{fig:maskratio}. After that, we compared the impact of different random mask ratio factors \(p\) on per-token loss reduction process, as illustrated in Figure~\ref{fig:loss}. From Figure~\ref{fig:maskratio} and Figure~\ref{fig:loss}, we can draw the following three conclusions:

\begin{enumerate}[(i)]
        \item In the early stages of training (less than 400 steps), a higher \(p\) value results in greater per-token loss. In the later stages (after 500 steps), the per-token loss converges regardless of the value of \(p\). This phenomenon is intuitive as the mask ratio factor effectively amplifies the weight of the diff code tokens loss on pre-trained LLM, the loss of diff code is greater than the loss of consistent code that has appeared in the prompt. As the model gradually converges, the difference in per-token loss between the two diminishes.
        \item Generally, the higher value of p, the fewer training steps are required to reach the checkpoint with the best bug-fixing capability. This is a key advantage of dynamic mask SFT, in addition to its ability to enhance the model's bug-fixing capabilities. This allows for improved model performance with lower computational costs and energy consumption. 
        \item From Figure~\ref{fig:maskratio}, we can clearly see that when the value of \(p\) is between $[0.4, 0.7]$, all the trained models achieve optimal performance. When the value of \(p\) is 1 (completely ignoring the loss of identical code lines), the performance of the model is worse than those using the default generative SFT (where p is 0). 
\end{enumerate}

The manual evaluation results of the ablation experiments shown in Table~\ref{tab:masksftexp} have adequately demonstrated the effectiveness and applicability of the Progressive Dataset Construction (PDC) data collection method and the Dynamic Mask SFT (DM-SFT) training approach in enhancing the LLM's capability for SQL code bug fixing. It is noteworthy that by setting parameter \(p\) appropriately, the dynamic mask SFT method can enhance the model's bug fixing capability while significantly reducing the training time. This allows the model to achieve optimal performance at earlier training steps. This is appealing given the high cost of computational resources of LLM training. In later model iterations, we experimented with using more training data, and DM-SFT maintained its advantage. This detailed in Appendix~\ref{sec:data_scaleup}.

\section{Conclusion}

In this paper, we innovatively propose a set of methods to enhance LLMs for SQL bug fixing, from data construction and model training aspects. For data construction, we introduce two approaches: a breadth-first diverse collecting method and a depth-first oriented generation method. The diverse collecting method mines user behavior for annotated data reflecting real-world scenarios distribution. The oriented generation method targets specific model weaknesses by data augmentation. Both methods are sustainable iteration and semi-automated, requiring minimal manual labor. That's why named as Progressive Dataset Construction (PDC). For training methodology, we propose the dynamic mask SFT, which is applicable to all generative code bug repair tasks. This method improves bug-fixing capability by nearly $10\%$ compared to default SFT and reduces the training time.

\clearpage
\section*{Limitations}

\textbf{Only generate the modification code lines} We attempted a highly efficient and intuitively appealing approach that involves generating only the correct code for the diff sections. Specifically, our approach required the model to output the lines of code that needed modification and the corrected code after changes. This definition could handle all code rewriting operations, including additions (where a single line of original code is replaced by multiple lines), deletions (where multiple lines of original code are replaced by an empty string), and modifications (where multiple lines of original code are replaced by multiple lines of new code). Unfortunately, this method resulted in impaired model performance due to the lack of context in the outputs, making it challenging to achieve the accuracy of generating complete code, both in prompt engineering experiments on GPT-4 \citep{achiam2023gpt} and in SFT training on open-source code LLMs.

\noindent\textbf{Token level dynamic mask SFT} A pertinent question arises as to why consistent lines cannot use token-level dynamic masking and must instead be masked by code lines. Indeed, in our earliest practices, we masked at the token level. However, perplexingly, models masked at the token level struggled to converge, and during evaluations, a portion of the samples consistently failed to generate complete and usable code. This remains a puzzle we have not fully resolved. We hypothesize that for programming languages, a line may correspond to a more complete semantic module, and token-level masking disrupts this contextual integrity. Research on this aspect will continue in subsequent studies. 

\bibliography{custom}

\newpage
\appendix

\onecolumn

\section{Appendix}
\label{sec:appendix}

\subsection{Bug SQL generation prompt of oriented generation method}
\label{sec:appprompt}

\begin{tcolorbox}[colback=blue!5!white, colframe=blue!75!black, title=Prompt]
Based on the SCHEMAS and TARGET SQL, help to generate the error \texttt{sql} which are related to SCHEMAS and similar to TARGET SQL. The generated error \texttt{sql} should contain error related to ERROR INFO. You should obey the following RULES.

\textbf{RULES}
\begin{enumerate}
    \item If the SCHEMAS are empty, it means the TARGET SPARK SQL is not related to any schemas.
    \item ERROR INFO should not be appeared in explanation.
    \item Except for error part of code, other parts of code should be same between correct \texttt{sql} and error \texttt{sql}.
    \item Comments and indents in generated error \texttt{sql} and correct \texttt{sql} should be the same.
    \item If it is hard to generate error \texttt{sql} which is similar to the TARGET SQL related to ERROR INFO, please return no in suitable field, otherwise it should be yes.
\end{enumerate}

Below is a brief example which you can refer to (if the slots of example is empty please ignore Example section):

\textbf{[EXAMPLE]}

\textbf{target \texttt{sql}:}

TARGET\_SQL\_EXAMPLE\_PLACEHOLDER

\textbf{error info:}

ERROR\_INFO\_EXAMPLE\_PLACEHOLDER

\textbf{error \texttt{sql}:}

ERROR\_SQL\_EXAMPLE\_PLACEHOLDER

Now give you the tables schema, corresponding target SQL and error type information as below.

Please write a error SQL that match the error type information.

\textbf{[SCHEMAS]}

SCHEMAS\_PLACEHOLDER

\textbf{[TARGET SPARK SQL]}

TARGET\_SPARK\_SQL\_PLACEHOLDER

\textbf{[ERROR INFO]}

ERROR\_INFO\_PLACEHOLDER

\textbf{RESPONSE REQUIREMENT}

Return \texttt{json} str which can be parsed by \texttt{json.loads()} of python3 as following:

\{\texttt{"error sql": "", "correct sql": "", "reason": "", "suitable": ""}\}

\end{tcolorbox}

\newpage
\subsection{Bug fixing model's input prompt}
\label{sec:online_prompt}
\begin{tcolorbox}[colback=blue!5!white, colframe=blue!75!black, title=Prompt]
Requirements: Directly generate the right SQL.

\textbf{[TABLES SCHEMA]}

TABLES\_SCHEMA\_PLACEHOLDER

\textbf{[BUG SQL]}

BUG\_SQL\_PLACEHOLDER

\textbf{[ERROR MESSAGE]}

ERROR\_MESSAGE\_PLACEHOLDER

\textbf{Question: BUGFIX task}

Based on the error SQL code, error messages, and input table schema, please fix the bugs and write the corresponding correct SQL code. Remember not to change any existing comments and SQL code without errors.

\textbf{Response:}

\end{tcolorbox}

\subsection{Experiments of Fine-Tuning Dataset Scaling Up}
\label{sec:data_scaleup}

We progressively collected larger training datasets using the PDC method and continuously conducted experiments on the effectiveness of dynamic mask SFT(DM-SFT) and default SFT(SFT). After model deployed in production environment, we expand the dataset every two months. So far, experiments have been conducted on SFT datasets of sizes 6.5k, 9.2k, and 12k. To maintain consistency, the evaluation still uses the 1072-size dataset mentioned earlier (the larger dataset is extensions of the smaller one).

\begin{enumerate}[(1)]
        \item \textbf{6.5k dataset: }5.9k of diverse collecting, 0.6k of oriented generation.
        \item \textbf{9.2k dataset: }7.6k of diverse collecting, 1.6k of oriented generation. 
        \item \textbf{12k dataset: }9.6k of diverse collecting, 2.4k of oriented generation. 
\end{enumerate}

Table~\ref{tab:scaleup} shows the human evaluation accuracy on the test dataset(1072) using DeepSeek-Coder-instruct-6.7B and CodeQwen1.5-Chat-7B as base models, comparing the DM-SFT method with the default SFT method across various training dataset sizes. It is clear that even as more training data is collected, the DM-SFT method can consistently maintains its competitive edge.\\


\begin{table}[H]
  \centering
  \begin{tabular}{llccc}
    \hline
    \textbf{Model} & \textbf{Train set} & \textbf{Method} & \textbf{Acc} & \textbf{Acc improvement} \\ \hline
    \multirow{6}{*}{DeepSeek-Coder-instruct-6.7B} 
    & \multirow{2}{*}{6.6k} & SFT & 48.4\% & \multirow{2}{*}{\centering +5.0\%} \\ \cline{3-4}
    &  & DM-SFT & 53.4\% & \\ \cline{2-5}
    & \multirow{2}{*}{9.2k} & SFT & 51.2\% & \multirow{2}{*}{\centering +4.7\%} \\ \cline{3-4}
    &  & DM-SFT & 55.9\% & \\ \cline{2-5}
    & \multirow{2}{*}{12k} & SFT & 54.0\% & \multirow{2}{*}{\centering +4.9\%} \\ \cline{3-4}
    &  & DM-SFT & 58.9\% & \\ \hline
    \multirow{6}{*}{CodeQwen1.5-Chat-7B} 
    & \multirow{2}{*}{6.6k} & SFT & 48.3\% & \multirow{2}{*}{\centering +2.8\%} \\ \cline{3-4}
    &  & DM-SFT & 51.1\% & \\ \cline{2-5}
    & \multirow{2}{*}{9.2k} & SFT & 51.9\% & \multirow{2}{*}{\centering +3.8\%} \\ \cline{3-4}
    &  & DM-SFT & 55.7\% & \\ \cline{2-5}
    & \multirow{2}{*}{12k} & SFT & 54.9\% & \multirow{2}{*}{\centering +3.8\%} \\ \cline{3-4}
    &  & DM-SFT & 58.7\% & \\ \hline
  \end{tabular}
  \caption{Accuracy of DM-SFT/SFT Across Various Size of Train Dataset.}
  \label{tab:scaleup}
\end{table}

\clearpage

\newpage
\subsection{Continue Pre-train}
\label{sec:cpt}



Throughout the model development phase, we compared the bug fixing capabilities of DeepSeek-Coder-6.7B-instruct and our internal code LLM on a case-by-case basis after fine-tuning them on the same dataset. We found that compared to the internal code LLM, DeepSeek-Coder is more prone to producing hallucination outputs when generating correct SQL code. Appendix~\ref{sec:appfigures} Figure~\ref{fig:hallucination} presents a typical example, where the left side shows the correct code snippet predicted by the internal code LLM (SFT), and the right side shows the correct code snippet predicted by the DeepSeek-Coder (SFT) model. The constant value $90000000$ of the original code was erroneously increased by an additional $0$ in DeepSeek-Coder model's prediction.

Through the analysis, we discovered that the differences in performance between the two foundation models which have been fine-tuned with the same supervised data may stem from their familiarity for the domain-specific SQL code style and distribution (the internal model’s pre-train corpus includes internal code data). To validate this hypothesis, we have mined, cleaned, and deduplicated a dataset from internal scenarios, and obtain a SQL code corpus with size of 53k. To ensure the rigor of the experiment, we carefully inspected these entries to guarantee that there would be no overlap with the 1072 samples in evaluation dataset. 

We conduct continue pre-train (CPT) \citep{ke2023continual} on the 53k domain-specific corpus which we have cleaned and use DeepSeek-Coder-6.7B-instruct, DeepSeek-Coder-V2-Lite-Instruct and CodeQwen1.5-7B-Chat as foundation models. We then compared the capabilities of the models with and without continued pre-training, As illustrated in Appendix~\ref{sec:appfigures} Figure~\ref{fig:ctperf}. We made some adjustments to the learning rate, setting it to $1.5e-5$ for continue pre-train and later tune it to $1.0e-5$ for subsequent SFT/DM-SFT. Through comparison, it is evident that after continue pre-train with domain-specific data, the 6 combinations of models and training methods achieved a bug-fixing accuracy improvement range of $1.3\% \sim 2.3\%$. Additionally, the number of bad cases which involved with hallucination modification has decreased across all models. 

There's worth mentioning that when using different models for continue pre-train, we adhered to the same input formats as their original pre-train. Additionally, we compared two training methods: training all parameters versus training the parameters outside of the embedding layer only during continue pre-train. Although the parameters of the embedding layer constitute only a small portion of the total parameters in most LLMs (for example, in DeepSeek-Coder 6.7b, the embedding layer accounts for approximately $1.96\%$ of whole parameters), training with the embedding layer parameters frozen has proven challenging to achieve the expected results in our practice. In Appendix~\ref{sec:appfigures} Figure~\ref{appendix:pre-train_loss}, we have documented the training loss decline curves for both full parameter continue pre-train and continue pre-train with only non-embedding layer parameters updated. It is evident that training with only non-embedding layer parameters updated struggles to converge, whereas full parameter update in continue pre-train demonstrates good convergence.

Finally, all source code related to our experiments are made publicly available in the corresponding GitHub repository\footnote{https://github.com/D1026/sql-bugfix-public}. Except for continue pre-train data, all other SFT data and evaluation dataset released in the same repository after anonymization. The continue pre-train data is included in another SQL corpus opening initiative and is not currently available separately.

\clearpage

\newpage
\subsection{Figures}
\label{sec:appfigures}

\begin{figure*}[ht]
\centering
\includegraphics[width=1.0\textwidth]{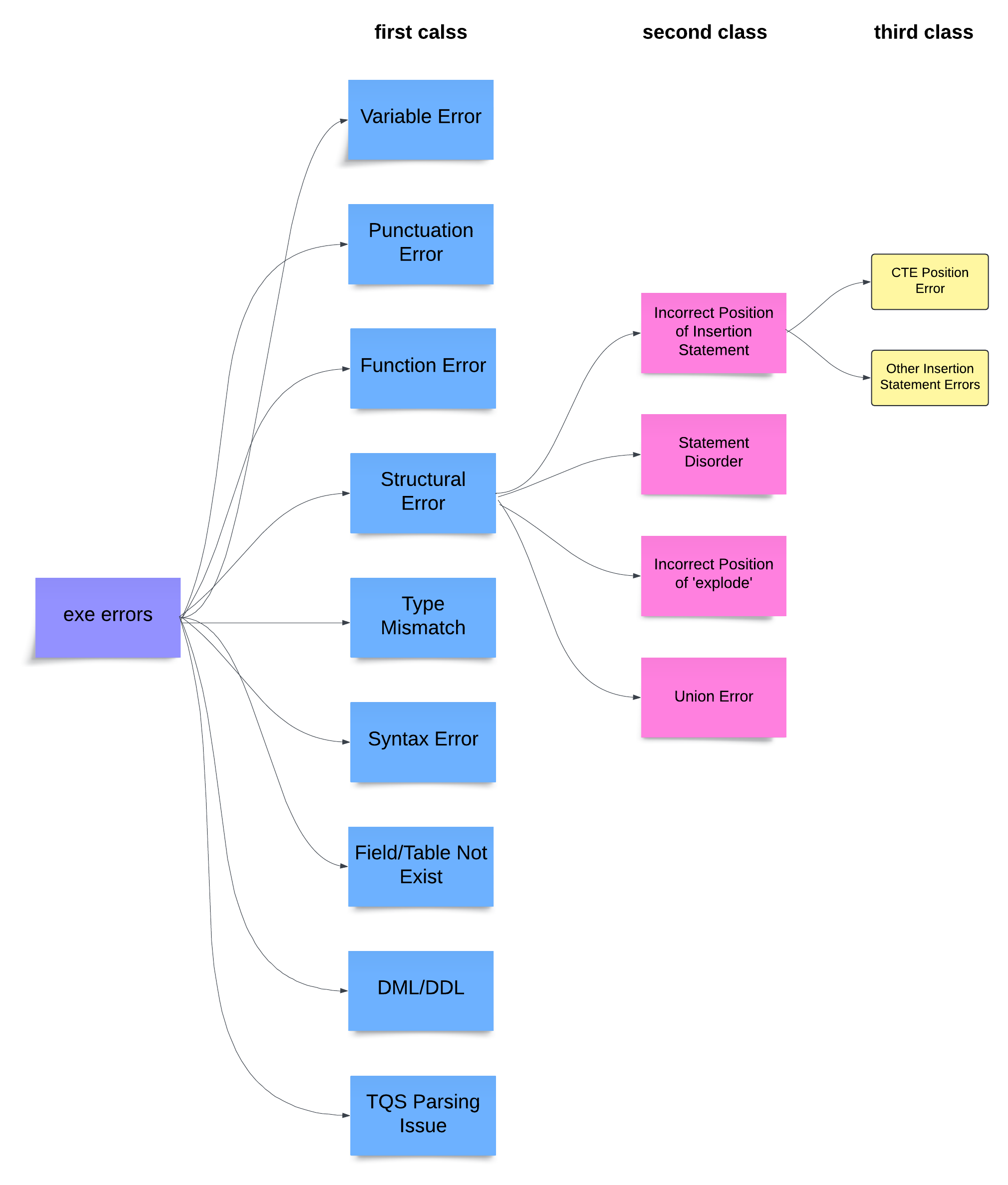}
\caption{SQL bugs three level classification}
\label{appendix:bug_types}
\end{figure*}

\noindent\textbf{Figure~\ref{appendix:bug_types}} illustrates SQL bugs categorized into a three-level classification by using an automated method based on error messages and SQL code, ultimately classifying all resolvable errors into 81 subcategories. \\

\begin{figure}[t]
\includegraphics[width=\columnwidth]{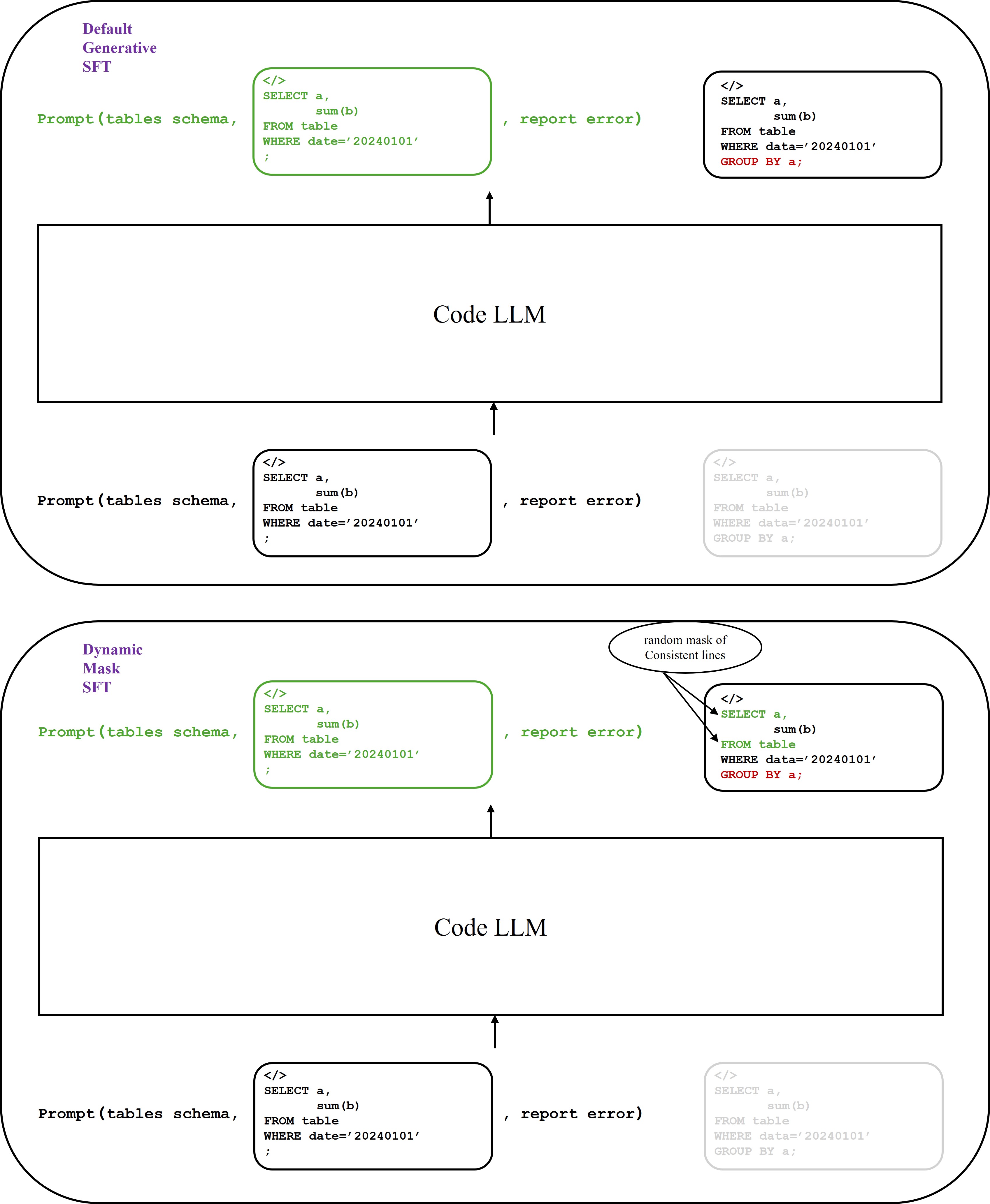}
\caption{A comparison of the default generative SFT (top) and dynamic mask SFT (bottom) for the code bug-fixing task.}
\label{fig:masksft-bigger}
\end{figure}

\noindent\textbf{Figure~\ref{fig:masksft-bigger}} is a larger and clearer version of Figure~\ref{fig:masksft}. \\

\begin{figure*}[ht]
\centering
\includegraphics[width=1.0\textwidth]{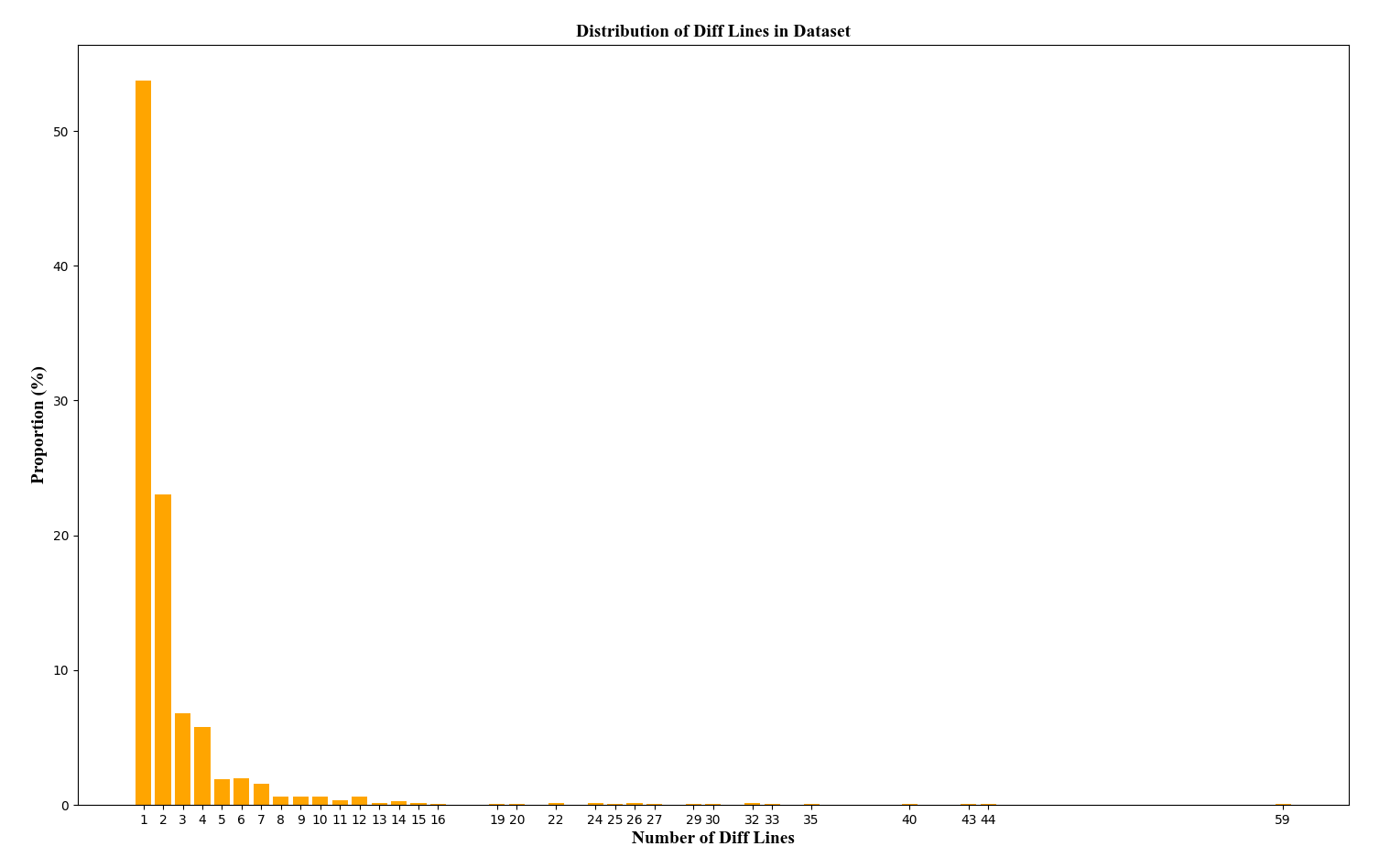}
\caption{Distribution of diff lines proportion in SQL code}
\label{appendix:difflines}
\end{figure*}

\noindent\textbf{Figure~\ref{appendix:difflines}} illustrates the distribution of the number of diff code lines in our collected training data. It can be observed that over $50\%$ of the bug SQL code require only a single line modification to be transformed into correct SQL code. \\

\begin{figure*}[ht]
  \includegraphics[width=\columnwidth]{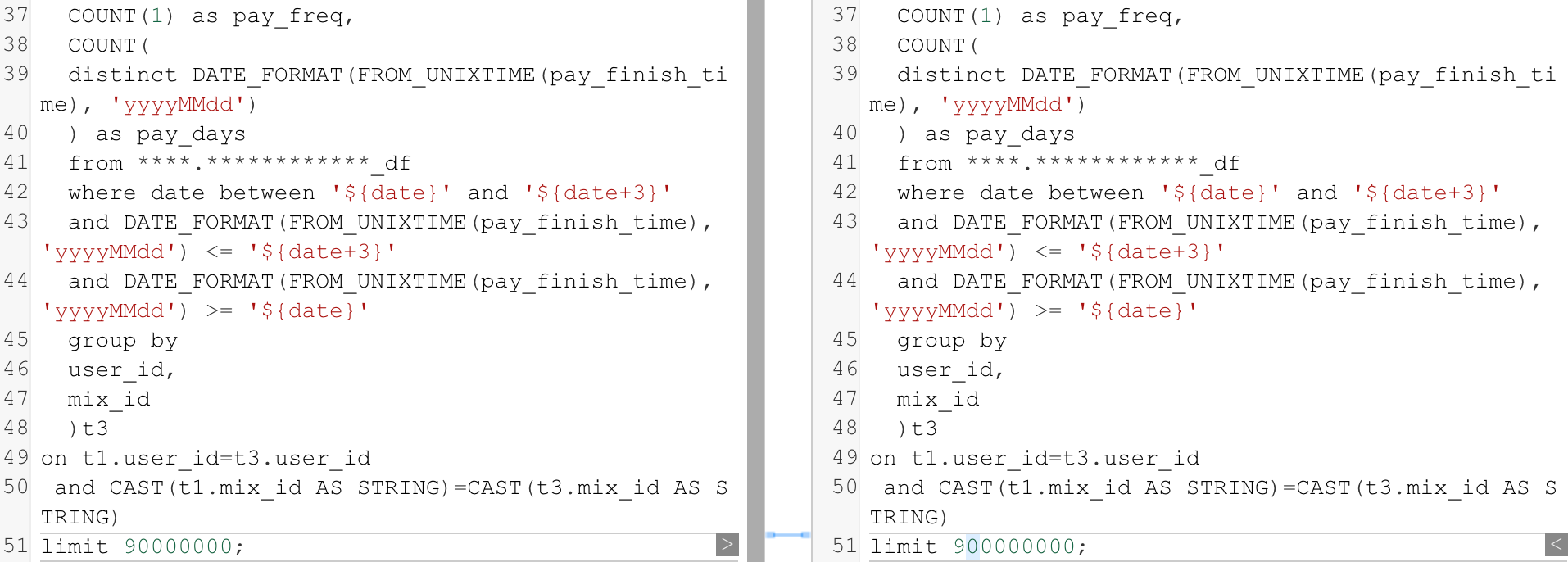}
  \caption{Hallucination modification by DeepSeek-Coder. Left: Output from internal code LLM (limit value consistent with original code). Right: Output from DeepSeek-Coder-Bugfix (limit value erroneously increased by an additional $0$ character).}
  \label{fig:hallucination}
\end{figure*}

\noindent\textbf{Figure~\ref{fig:hallucination}} illustrates a typical case where the internal code LLM successfully maintains the constant '90000000'. Meanwhile, the code generated by the DeepSeek incorrectly adds an extra '0' to the constant '90000000'. Although both two model have trained by same SFT dataset. Our internal code LLM pre-train corpus includes a substantial amount of internal SQL code. In comparison, the proportion of SQL code in DeepSeek's pre-train data is minimal. This may lead to the differences. \\

\begin{figure}[t]
  \includegraphics[width=\columnwidth]{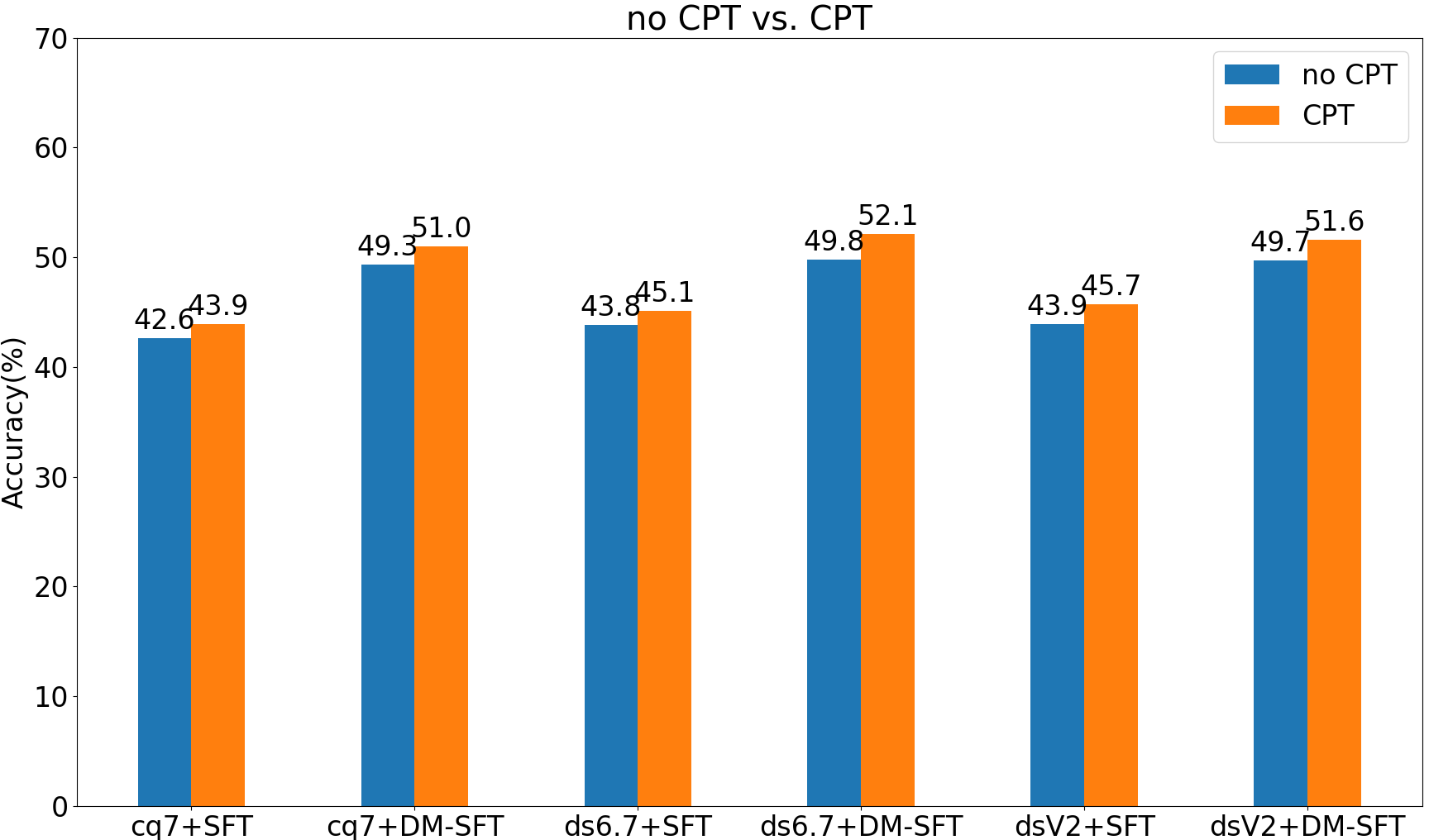}
  \caption{ Performance differences of models with and without continued Pre-train on domain-specific corpus (pre-train before SFT/DM-SFT). 'cq7':CodeQwen1.5-7B-Chat, 'ds6.7':DeepSeek-Coder-6.7B-instruct, 'dsV2':DeepSeek-Coder-V2-Lite-Instruct}
  \label{fig:ctperf}
\end{figure}

\noindent\textbf{Figure~\ref{fig:ctperf}} presents the bug-fixing accuracy differences across six combinations: three models X two SFT training methods, with and without continued pre-train of the base model. \\

\begin{figure*}[ht]
\centering
\includegraphics[width=1.0\textwidth]{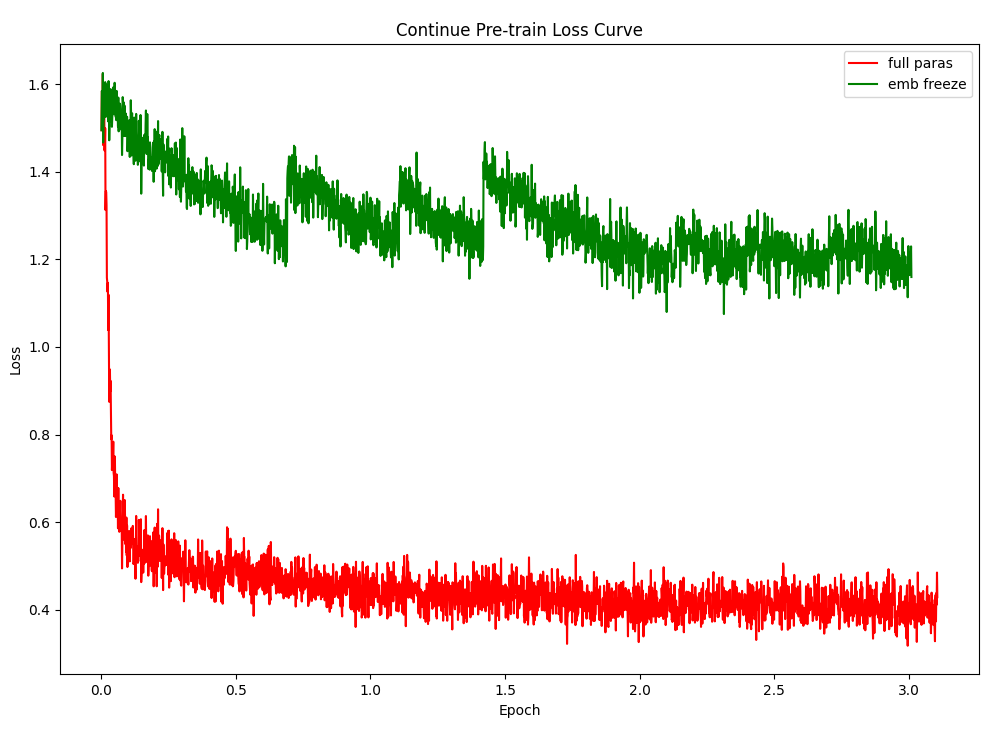}
\caption{Training Loss Curve for Two Continue Pre-train Methods}
\label{appendix:pre-train_loss}
\end{figure*}

\noindent\textbf{Figure~\ref{appendix:pre-train_loss}} clearly demonstrates the differences in loss reduction when performing continued pre-train on in-domain SQL code corpus, comparing full-parameter training and training with frozen embedding layer parameters. Despite the embedding layer parameters constituting less than $2\%$ of the total parameters in DeepSeek-Coder6.7b, the loss reduction during continue pre-train with frozen embedding layer parameters is highly unstable. Moreover, the final converged loss value shows a significant disparity compared to full-parameter continue pre-train. 

\clearpage

\end{document}